\documentclass{article}
\usepackage{spconf,amsmath,graphicx}
\usepackage{xcolor}
\usepackage{cite}
\usepackage{todonotes}
\usepackage{hyperref}
\usepackage{multirow}
\usepackage{bm}
\usepackage{todonotes}
\usepackage[tone,extra,safe]{tipa}
\usepackage{booktabs}
\usepackage{amssymb}
\usepackage{tipa}

\title{Memory Augmented Lookup Dictionary based Language Modeling for Automatic Speech Recognition}

\name{\begin{tabular}{c}{}Yukun Feng$^{1,*}$\thanks{$^*$Work done during internship at ByteDance},
Ming Tu$^{2,\dagger}$\thanks{$^{\dagger}$Corresponding author}, Rui Xia$^2$, Chuanzeng Huang$^2$, Yuxuan Wang$^2$\end{tabular}
}

\address{
Johns Hopkins University$^1$ \\
Speech and Music Intelligence (SAMI), ByteDance$^2$\\
\begin{footnotesize}
\texttt{yfeng55@jhu.edu, \{mingtu, rui.xia, huangchuanzeng, wangyuxuan.11\}@bytedance.com}
\end{footnotesize}
    }
\begin{document}
\ninept
%
\maketitle
\begin{abstract}
Recent studies have shown that using an external Language Model (LM) benefits the end-to-end Automatic Speech Recognition (ASR).
However, predicting tokens that appear less frequently in the training set is still quite challenging.
The long-tail prediction problems have been widely studied in many applications, but only been addressed by a few studies for ASR and LMs.
In this paper, we propose a new memory augmented lookup dictionary based Transformer architecture for LM. The newly introduced lookup dictionary incorporates rich contextual information in training set, which is vital to correctly predict long-tail tokens. With intensive experiments on Chinese and English data sets, our proposed method is proved to outperform the baseline Transformer LM by a great margin on both word/character error rate and tail tokens error rate. This is achieved without impact on the decoding efficiency.
Overall, we demonstrate the effectiveness of our proposed method in boosting the ASR decoding performance, especially for long-tail tokens.

\end{abstract}

%


\section{Introduction}
\label{sec:intro}
While a lot of studies have demonstrated the superiority of end-to-end (E2E) Automatic Speech Recognition (ASR) systems \cite{graves2012sequence,chan2015listen} and the effectiveness of incorporating Language Models (LM) into the E2E ASR systems\cite{toshniwal2018comparison,kannan2018analysis}, recognition and prediction of words that appear only a few or zero times in training data are still big challenges, especially for E2E ASR systems which are optimized only on text in  the training data. 

Some studies have addressed this long-tail problem for E2E ASR \cite{peyser2020improving1,peyser2020improving2,winata2021adaptandadjust,9664007,Huang2022SentenceSelectLL,Huang2021LookupTableRL,yang2021multi}.
The studies in \cite{peyser2020improving2,Huang2022SentenceSelectLL} resort to adding large corpora of textual data or adjusting the distribution of head and tail words in LM training to improve the modeling ability of tail words. In \cite{winata2021adaptandadjust,9664007}, the authors propsoed to improve the prediction of tail words with the help of large-scale pretrained LMs (BERT\cite{devlin-etal-2019-bert} variants) which inevitably increases the decoding computational cost. Another line of research modified the training loss or introduced extra loss terms to regularize the ASR training, and results showed improved performance on rare words \cite{peyser2020improving1,yang2021multi}.
In \cite{Huang2021LookupTableRL}, the authors tried to scale up the embedding capacity of an RNN LM by incorporating N-gram context embedding into the embedding layer without sacrificing decoding efficiency.
However, it ignored the frequency information of words and N-grams and only reply on the input embedding layer to learn enough contextual information to predict rare words.

Since Transformer LMs have shown better performance than RNN LMs for ASR \cite{Irie2019LanguageMW}, in this paper we extend the Transformer LMs with 
a lookup dictionary that maps the current context to candidate tokens that have occurred during training. Inspired by \cite{taking_notes} which focuses on effective training of BERT,
we initialize a dictionary by aggregating the N-gram token IDs of the current token as keys and utilize a multi-vector array as values to enable memorization of rich context information.
We now consider the dictionary's values as the memory of the corresponding N-gram context.
Specifically, the contextual memory is updated by the current token's \textbf{subsequent} token embedding in the training based on how often the \textbf{subsequent} token occurs in the training corpus.
For each key, the frequency of the \textbf{subsequent} token decides how many vectors in the corresponding multi-vector value will be updated.
We then use an attention module at the last layer of the transformer blocks to map the dictionary memory to the contextualized embedding of the current token, in which the current context will query the most relevant vectors from the corresponding multi-vector memory.

\begin{figure*}[t]
\includegraphics[scale=0.4]{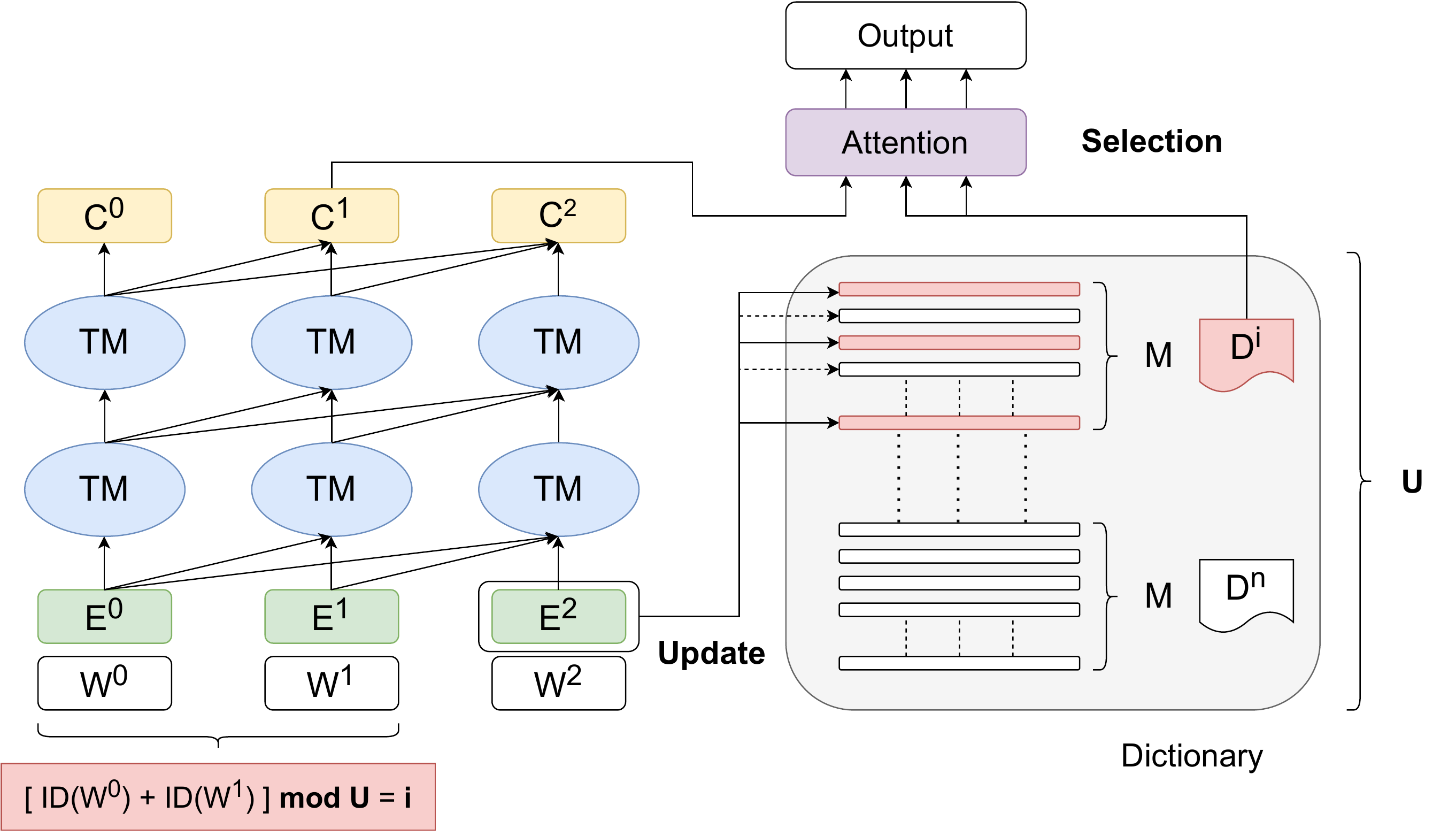}
\centering
\caption{Overview of the proposed memory augmented lookup dictionary based Transformer LM. $W^k$ represents the input tokens, $E^k$ is for the token embedding,"TM" means Transformer blocks in auto-regressive manner. $C^k$ is the contextualized embedding corresponding to the input tokens. Input and output embedding weights are shared in the LM.}
\label{fig-model}
\end{figure*}

We experimented on two Mandarin ASR data sets and improve 8.5\% relatively of the Character Error Rate (CER) over the baseline Transformer LM.
Notably, our method show 13\% and 12.5\% relative CER reduction on the 1-gram and 2-gram tail tokens.
Also, we achieve Word Error Rate (WER) improvement on the two test sets of LibriSpeech.
The results indicate the success of our method on improving not only the general ASR decoding but also the prediction of tail tokens for both Mandarin and English.
We did intensive analysis to
investigate the benefits of different aspects of our proposed method.
Overall, this paper makes several contributions as the following:
\begin{enumerate}
\itemsep0em 

    \item We propose a new Transformer based LM for ASR equipped with a lookup dictionary consisting of multi-vector memory that builds bonding between the current context and to-be-predicted candidate tokens.
    \item We incorporate the N-gram context information and the token frequency in training data into the lookup dictionary to improve the prediction of rare words.
    \item Our proposed LM significantly outperforms the baseline LM in both Mandarin and English ASR while keeping the same inference efficiency.
\end{enumerate}

\section{Proposed approach} 
In \autoref{fig-model}, we show the diagram of the Transformer LM equipped with our proposed memory augmented lookup dictionary. Each module will be introduced separately in the following subsections.

\subsection{Dictionary Construction and Indexing}
We initialize the dictionary as $\pmb{\mathrm{D}} \in \mathbb{R}^\mathrm{{U \times M \times d_{emb}}}$, where $\mathrm{d_{emb}}$ is the embedding size and $\mathrm{U}$ is the dictionary size.
Instead of using one vector as value for each key as in \cite{taking_notes}, 
we scale up the dictionary size by introducing an extra hyper-parameter $M$ to form a key-value pair as $(i, \mathbf{D}^i)$ where $\mathbf{D}^i \in \mathbb{R}^\mathrm{ M \times d_{emb}}$. For the $k^{th}$ token in the input sequence, the corresponding dictionary index $i$ is mapped through a modular hash with $U$, defined as:
\begin{equation}
  i = {\mathrm{ID(Token_{k})}}~\mathrm{mod~U }
\label{eq-index}
\end{equation} 
where ID() refers to the vocabulary id of the input token.
We believe with multiple vectors stored for each entry, much richer contextual information could be memorized compared than the  single vector counterpart. To consider more context in dictionary indexing, we also extend \autoref{eq-index} to N-gram case as in \cite{Huang2021LookupTableRL}, where the dictionary index $i$ is calculated as follows
\begin{equation}
i = (\sum_{n = k-N+1}^k {\mathrm{ID(Token_{n})}})~\mathrm{mod~U }
\end{equation}
where $N$ indicates the number of token IDs to aggregate. For example, if $N=2$, we sum up the IDs of the current token and its previous token before the modulo operation.

To trade off information redundancy and memory capacity, collision is allowed when doing hashing. $U$, $N$ and $M$ can be adjusted appropriately. We assume this approach could utilize the dictionary memory more efficiently. We will show the influence of changing the three hyper-parameters to the performance in results part.

\subsection{Dictionary Update}
As the $k_{th}$ token is mapped to the dictionary memory $\mathbf{D}^{i}$ through \autoref{eq-index},
each memory vector $\mathbf{d}_{i}^m$ is updated by the embedding of the embedding of current token's next token $\mathbf{e}_{k+1}$, which can be formulated as:
\begin{equation}
  \widetilde{\mathbf{d}_{i}^m} =
  \begin{cases}
   \mathbf{d}_{i}^m*\alpha + \mathbf{e}_{k+1}*(1-\alpha) & \text{if $X_{k+1}$ = 1} \\
   \mathbf{d}_{i}^m & \text{if $X_{k+1}$ = 0}
  \end{cases}
\end{equation} 
where $\alpha$ is a smoothing hyper-parameter that indicates how much information comes from $\mathbf{e}_{k+1}$, and we set it as 0.5 for all later experiments.
We define a Bernoulli Variable $\mathrm{X_{k+1} \sim Bern({P_{k+1}})}$, which decides how many vectors will be updated in the matrix $\mathbf{D}^{i}$.
$P_{k+1}$ indicates the update ratio, computed by the normalized occurrence of the ${k+1}^{th}$ token in training data:
\begin{equation}
P_{k+1} = \frac{1}{\log{\text{(Count of } \mathrm{Token_{k+1}})}}.
\end{equation}
In this case, embeddings of low frequency tokens are able to contribute more to the corresponding memory compared to high frequency tokens. Token frequencies are calcualted with training text and corresponding text tokenizer.
We will further discuss the effect of the update ratio in \autoref{analysis-prob}.

\begin{table*}
\centering
\scalebox{0.9}{
\begin{tabular}{l|ccc|ccc}
\hline
\textbf{Model} & \multicolumn{3}{c}{\textbf{Input}} & \multicolumn{3}{c}{\textbf{Search}} \\ 
\hline
& Overall & Tail-1 & Tail-2 & Overall & Tail-1 & Tail-2 \\
& CER$/$SER & CER & CER & CER$/$SER & CER & CER \\
\hline
Conformer & 6.19 $/$ 44.91 & 15.87 & 13.82 & 13.51 $/$ 45.38 & 22.38 & 21.46  \\
\textbf{with Language Model} &  &  &  &  & \\
~+ LM & 5.55 $/$ 41.80 & 13.46 & 11.86 & 9.07 $/$ 30.11 & 13.86 & 13.78 \\
~+ LM~$\text{\cite{Huang2021LookupTableRL}}$ & 5.42 $/$ 40.37 & 12.50 & 11.18 & 8.91 $/$ 29.79 & 13.50 & 13.44 \\
~+ LM~$\text{\cite{taking_notes}}$ & 5.35 $/$ 40.19 & 12.48 & 10.97 & 9.02 $/$ 30.23 & 13.80 & 13.68 \\
~+ Ours & $\textbf{5.09}$ $/$ $\textbf{38.86}$ & $\textbf{11.46}$ & $\textbf{10.30}$ & $\textbf{8.29 $/$ 27.60}$ & $\textbf{12.34}$ & $\textbf{12.13}$ \\
 & +8.3$\%$ & +14.9$\%$ & +13.2$\%$ & +8.6$\%$ & +11.0$\%$ & +12.0$\%$ \\
\hline
~+ $LM_{L}$ & 4.83 $/$ 37.29 & 10.94 & 9.67 & 8.24 $/$ 27.21 & 12.26 & 12.27 \\
~+ $Ours_{L}$ & $\textbf{4.73}$ $/$ $\textbf{36.90}$ & $\textbf{10.54}$ & $\textbf{9.54}$ & $\textbf{7.80 $/$ 26.20}$ & $\textbf{11.49}$ & $\textbf{11.41}$ \\
 & +2.1$\%$ & +3.7$\%$ & +1.3$\%$ & +5.4$\%$ & +6.3$\%$ & +7.0$\%$ \\
\hline
\end{tabular}}
\caption{Evaluation of CER and SER on two internal Chinese ASR test sets. ``Input'' and ``Search'' refer to voice input and voice search domain test sets respectively. $L$ refers to the LM with 1024 embedding size.}
\label{zh-asr-results}
\end{table*}

\subsection{Context Selection}
We use an attention module to relate the output representation of the current token to the corresponding dictionary memory. 
Attention performs as a mapping function for the input query (Q) and key-value (K-V) pairs, as
\begin{equation}
    \mathrm{Attention}(\mathbf{Q},\mathbf{K},\mathbf{V}) = Softmax(\frac{\mathbf{Q}\mathbf{K}^T}{\sqrt{d_{emb}}})\mathbf{V}.
\end{equation}
As we share the input and output embedding weight in the Transformer LM, and the dictionary memory stores the candidate tokens' embedding during training, 
we assume the attention module could help select useful information from the memory given the output representation of the current token.
For the $k_{th}$ token, we define the contextualized token embedding from the Transformer model as $\mathbf{c}_{k}$, and its corresponding dictionary memory is defined as $\mathbf{D}^{i}$, where $i$ is the hashing index from Eq 1 or 2.
The new output representation $\widetilde{\mathbf{c}_k}$ is computed as:
\begin{equation}
    \widetilde{\mathbf{c}_k} = \mathrm{Attention}(\mathbf{c}_k, \mathbf{D}^{i}, \mathbf{D}^{i}),
\end{equation}
which will then be used to calculate the output token distribution.

\subsection{Training and Inference}
During training, the context selection operation was done before the dictionary update for the reason that the update information in Eq 3 for the current input sentence will not affect the current context selection.
To stabilize training, we also disable the dictionary update for the first 1000 training steps to warmup the newly initialized embedding to a good distribution.
During inference, the dictionary update is also disabled to avoid any information leakage for auto-regressive prediction. With the trained memory augmented Transformer LM, we apply shallow fusion to integrate the LM to ASR decoding with weight $\lambda_{sf}$. Also, Internal Language Model Estimation (ILME) \cite{Meng2021InternalLM} is adopted to suppress the internal LM of the E2E ASR and advocate the contribution of the external LM, which has been proved to be quite effective especially there is domain mismatch between textual distribution of ASR and LM training data. The weight of ILME is noted by $\lambda_{i}$. We also tried LM rescoring over the N-best output of beam search, and the weight of rescoring is noted by $\lambda_{res}$.

\section{Experiment}

\subsection{Datasets}
We adopt the LibriSpeech~\cite{7178964} dataset to evaluate the ASR performance in English.
We use the standard 960 hours data for training and the "clean" and "other" test sets for evaluation.
The corresponding LM is trained on PG-19~\cite{raecompressive2019}, an 11GB in-domain text corpus consisting of books extracted from Project Gutenberg.
To match the averaged sentence length in LibriSpeech, we process the PG-19 into a sentence-level corpus.
We use the unigram tokenizer \cite{kudo-2018-subword} with vocabulary size of 5000 from ESPnet \cite{watanabe18_interspeech} for both ASR and LM training.
Also, we evaluate our method on two internal Chinese video datasets.
We have a 10k hours annotated audio dataset for general ASR training and two test sets: one is voice input domain (5103 utterances) and the other is voice search domain (6424 utterances), which are two different domains compared to the ASR training set.
As for the LM training, we have a 60GB text corpus for the voice input domain and 2GB corpus for the voice search domain.
We process the Chinese text at the character level with a vocabulary size of 11k (with both Chinese characters and English subword tokens).

Besides evaluating the overall performance on the above mentioned test sets, we also assess the ASR metrics on tail tokens.
Tail tokens are defined as the tokens whose accumulated frequency in the training corpus is lower than a $threshold$, which we set as 5\%, i.e. the frequency ratio of head and tail tokens is 95:5. Both 1-gram (Tail-1) and 2-gram (Tail-2) tail tokens are extracted from test sets at character-level for Chinese. For English teset ses, we only extracted 1-gram word-level tail tokens.

\subsection{Experimental settings}
We train both Chinese and English ASR models with a LAS \cite{chan2015listen} architecture, for which we use a 12-layer Conformer\cite{inproceedings} encoder and 6-layer Transformer decoder for Librispeech (as in ESPnet), and a 18-layer Conformer encoder and 4-layer Transformer decoder for the 10k hours Chinese dataset.

For LibriSpeech, we configure the LM as a 16-layer Transformer blocks with 1024 embedding size (as in ESPnet).
It is trained on PG-19 for sentence-level language modeling with a dropout rate of 0.3 and an effective token number of 524288 in each update.
Adam with betas of (0.9, 0.98), and weight decay of 0.01 is used for the optimization with 10k warmup steps. For the proposed look-up dictionary, we use 2-gram for dictionary hashing (as in Eq. 2); $U$ is set to 5k; $M$ is set to 64. The LM for Chinese datasets consists of 4 layer Transformer blocks with the embedding size of 384 and 1024 for small and large configuration respectively \footnote{We avoided the LibriSpeech settings because of impact on the decoding efficiency.}. For look-up dictionary, $U$ is set to 10k and other hyper-parameters are the same with the Librispeech settings.

For ASR inference in this paper, we set $\lambda_{sf}$=\{0.15, 0.4, 0.4\}, $\lambda_{res}$=\{0.0, 0.0, 0.1\}, $\lambda_{i}$=\{0.0, 0.2, 0.2\} for \{"LibriSpeech",  "Input, "Search"\} respectively, which give the best performance. A beam size of 60 is used for the LibriSpeech and 10 for Chinese sets. We use Word Error Rate (WER) as ASR metric for LibriSpeech test sets, and
Character Error Rate (CER) and Sentence Error Rate (SER) for Chinese test sets. For both test sets, we also calculate the tail token error rate by only counting errors on tail tokens and ignoring errors on other tokens within the same testing utterances.

\subsection{Results}
We compare our model with the original Transformer LM, as well as two other baselines: N-gram augmented embedding for LM training in \cite{Huang2021LookupTableRL} and single-vector memory for BERT pretraining in \cite{taking_notes}.
In \autoref{zh-asr-results}, while the original LM helps the ASR model achieve lower CER and SER, our method shows significant improvement over it and the two baseline methods.
We achieve 8.3\% and 8.6\% CER improvement on the general "Input" and "Search" test sets over the Transformer LM, and the CER improvement of tail tokens are even higher: 13\% on 1gram and 12.5\% on 2gram tail tokens.
As we increase the hidden size of the LM from 384 to 1024, the performance gain is not as much as the small LMs, but our method still outperforms the LM by ~3.7\% on overall CER and 4.6\% on tail tokens CER. In \autoref{en-asr-results}, our proposed method also shows consistent improvement on the two LibriSpeech test sets. The improvement on tail word error rate is more significantly compared to the overall WER improvement as on Chinese test sets.

\begin{table}
\centering
\begin{tabular}{lcccc}
\hline
\textbf{Model} & \multicolumn{2}{c}{\textbf{Clean}} & \multicolumn{2}{c}{\textbf{Other}}\\ 
 & {Overall} & {Tail-1} & {Overall} & {Tail-1} \\ 
\hline
Conformer  & 3.12\% & 11.92\% & 6.23\% & 24.52\% \\
~+~LM  & 3.08\% & 10.93\% & 5.81\% & 23.30\% \\
~+~Ours & \textbf{3.01\%} & \textbf{10.57\%} &  \textbf{5.73\%} & \textbf{22.93\%} \\

\hline
\end{tabular}
\caption{Evaluation of WER on the LibriSpeech test sets.}
\label{en-asr-results}
\end{table}

\section{Analysis}
In this section, we analyze how the different hyper-parameters, including dictionary size $U$, $N$ in N-gram for hashing, memory update ratio and memory size of each entry $M$, affect the performance.
All experiments are conducted on the Chinese "Search" test set, and the Transformer LM model with the proposed memory augmented lookup dictionary has 4 layers and 384 hidden size.

\begin{figure}[t]
\includegraphics[scale=0.12]{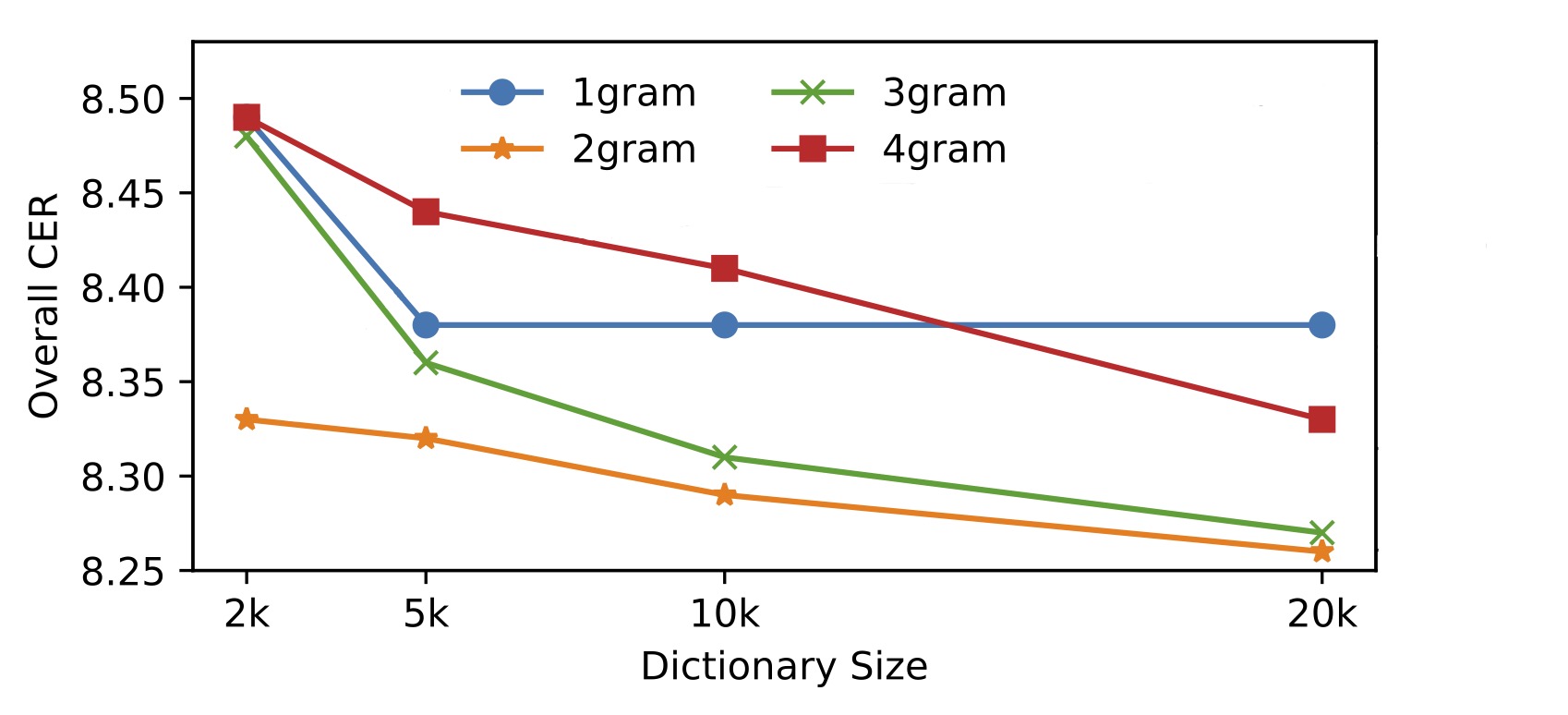}
\centering
\caption{Change of the overall CER (\%) on the "Search" test set with different dictionary size and different N-gram settings.}
\label{fig-ngram}
\end{figure}

In \autoref{fig-ngram}, we show the change of the overall CER (y axis) with the increase of dictionary size in different N-gram settings. It is clear that for each N-gram setting, increasing the dictionary size will boost the performance, and 2-gram achieves the best performance. Since the degree of collision elevates with bigger $U$ and $N$, larger $N$ means more collision; thus 4-gram performs even worse than 1 gram case when the dictionary size is not large. Considering the extra space taken by large dictionary size, wo choose the 2-gram with 10k dictionary size.

\begin{table}[t]
\centering
\begin{tabular}{lccc}
\hline
\textbf{Ratio} & \textbf{Overall} & \textbf{Tail-1}  & \textbf{Tail-2}\\ 
\hline
0.2 & 8.37\% & 12.47\% & 12.37\% \\
0.5 & 8.35\% & 12.38\% & 12.24\% \\
0.8 & 8.31\% & 12.34\% & 12.16\% \\
freq & 8.29\% & 12.34\% & 12.13\% \\
\hline
\end{tabular}
\caption{Change of the overall and tail tokens CER under different memory update options.}
\label{analysis-prob}
\end{table}


\begin{figure}[t]
\includegraphics[scale=0.48]{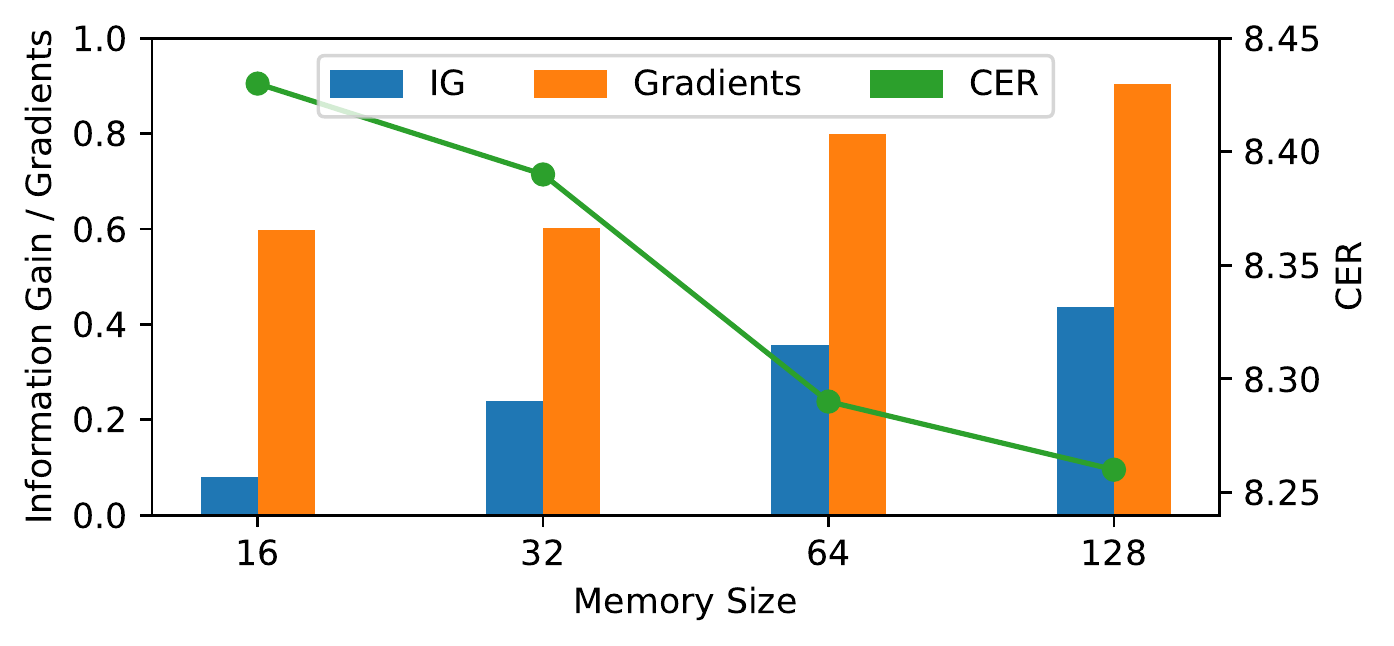}
\centering
\caption{
Overall CER, Gradients, and Information Gain (IG) change on the "Search" test set with the increase of memory size $M$}
\label{analysis-memory}
\end{figure}

In \autoref{analysis-prob}, we show the performance of both the overall CER and CER on tail tokens under different memory update settings. The ratios "0.2", "0.5" and "0.8" indicate we set a fixed probability for all tokens when sampling the Bernoulli variable $X_{k+1}$ in Eq. 3 to update the memory, while the "freq" means we use Eq. 4 to decide the $P_{k+1}$ for different tokens depending on their frequency in training set. The results demonstrate that large update ratio tends to improve the performance and our proposed frequency-based memory update strategy marginally beat other options.


\autoref{analysis-memory} analyzes if a large memory size $M$ would help the selection and the overall performance.
We use the Information Gain (IG) which is computed by the difference in the attention entropy (as in Eq. 5 and 6) between a randomly initialized dictionary and a well-trained one.
The entropy indicates how well the dictionary maps the information to the contextualized embedding of the current token $\widetilde{\mathbf{c}_k}$\cite{feng-etal-2022-learn}. 
The results show the IG is highly correlated with $M$. 
Besides, we adopt the Gradient Attribution test \cite{ancona2018towards,https://doi.org/10.48550/arxiv.2202.07856} to address the dictionary memory's contribution further. 
It computes the normalized gradient of the model variables to reflect its contribution to the output prediction.
It shows the gradients are also consistent with the previous finding that a larger memory would receive more gradients, indicating a greater contribution to the model prediction.  
However, considering the small relative gain and high computational cost when we increase the memory size from 64 to 128, we set the memory size as 64 in our experiments.

Finally, we want to discuss how the proposed memory augmented lookup dictionary will affect the model size and inference speed. During inference, compared to the baseline Transformer LM, the additional computation of our method is only the dictionary indexing (Eq. 2) and context selection (Eq. 6).
For lookup dictionary, the indexing operation requires $O(1)$ time cost.
The context selection also performs as a constant time cost as $O(M)$, where $M$ is the memory size of the dictionary.
We evaluate the Real Time Factor (RTF) on the "Search" test set on a NVIDIA A100 GPU with beam size batch size equals to 1.
The RTF is 0.124 for ASR model only, and 0.195 and 0.198 for the baseline Transformer LM and our proposed LM, respectively.
We notice that such additional operations almost do not affect the decoding speed in practice though the model size increases by introducing the lookup dictionary. 

\section{Conclusions}
In this paper, we propose a memory augmented lookup dictionary based Transformer LM to improve the language modeling in ASR, especially for long tail tokens.
We have improved the baseline Transformer LM in terms of overall ASR metrics and the tail words error rate in both Chinese and English test sets.
We also analyze our method under different hyper-parameter settings.
Overall, the results prove the superiority of the method over the baseline Transformer LM without sacrificing inference speed. Future work includes more experiments on English data sets, especially in domain mismatch condition. We are also interested in applying the method to general language modeling tasks.


\clearpage
\bibliographystyle{IEEEtran}
\begin{small}
\bibliography{refs,strings}

\begin{thebibliography}{10}
\providecommand{\url}[1]{#1}
\csname url@samestyle\endcsname
\providecommand{\newblock}{\relax}
\providecommand{\bibinfo}[2]{#2}
\providecommand{\BIBentrySTDinterwordspacing}{\spaceskip=0pt\relax}
\providecommand{\BIBentryALTinterwordstretchfactor}{4}
\providecommand{\BIBentryALTinterwordspacing}{\spaceskip=\fontdimen2\font plus
\BIBentryALTinterwordstretchfactor\fontdimen3\font minus
  \fontdimen4\font\relax}
\providecommand{\BIBforeignlanguage}[2]{{%
\expandafter\ifx\csname l@#1\endcsname\relax
\typeout{** WARNING: IEEEtran.bst: No hyphenation pattern has been}%
\typeout{** loaded for the language `#1'. Using the pattern for}%
\typeout{** the default language instead.}%
\else
\language=\csname l@#1\endcsname
\fi
#2}}
\providecommand{\BIBdecl}{\relax}
\BIBdecl

\bibitem{graves2012sequence}
A.~Graves, ``Sequence transduction with recurrent neural networks,''
  \emph{arXiv preprint arXiv:1211.3711}, 2012.

\bibitem{chan2015listen}
W.~Chan, N.~Jaitly, Q.~V. Le, and O.~Vinyals, ``Listen, attend and spell,''
  \emph{arXiv preprint arXiv:1508.01211}, 2015.

\bibitem{toshniwal2018comparison}
S.~Toshniwal, A.~Kannan, C.-C. Chiu, Y.~Wu, T.~N. Sainath, and K.~Livescu, ``A
  comparison of techniques for language model integration in encoder-decoder
  speech recognition,'' in \emph{2018 IEEE spoken language technology workshop
  (SLT)}.\hskip 1em plus 0.5em minus 0.4em\relax IEEE, 2018, pp. 369--375.

\bibitem{kannan2018analysis}
A.~Kannan, Y.~Wu, P.~Nguyen, T.~N. Sainath, Z.~Chen, and R.~Prabhavalkar, ``An
  analysis of incorporating an external language model into a
  sequence-to-sequence model,'' in \emph{2018 IEEE International Conference on
  Acoustics, Speech and Signal Processing (ICASSP)}.\hskip 1em plus 0.5em minus
  0.4em\relax IEEE, 2018, pp. 1--5828.

\bibitem{peyser2020improving1}
C.~Peyser, T.~N. Sainath, and G.~Pundak, ``Improving proper noun recognition in
  end-to-end asr by customization of the mwer loss criterion,'' in \emph{ICASSP
  2020-2020 IEEE International Conference on Acoustics, Speech and Signal
  Processing (ICASSP)}.\hskip 1em plus 0.5em minus 0.4em\relax IEEE, 2020, pp.
  7789--7793.

\bibitem{peyser2020improving2}
C.~Peyser, S.~Mavandadi, T.~N. Sainath, J.~Apfel, R.~Pang, and S.~Kumar,
  ``Improving tail performance of a deliberation e2e asr model using a large
  text corpus,'' \emph{arXiv preprint arXiv:2008.10491}, 2020.

\bibitem{winata2021adaptandadjust}
\BIBentryALTinterwordspacing
G.~I. Winata, G.~Wang, C.~Xiong, and S.~Hoi, ``Adapt-and-adjust: Overcoming the
  long-tail problem of multilingual speech recognition,'' 2021. [Online].
  Available: \url{https://openreview.net/forum?id=34KAZ9HbJco}
\BIBentrySTDinterwordspacing

\bibitem{9664007}
K.~Deng, G.~Cheng, R.~Yang, and Y.~Yan, ``Alleviating asr long-tailed problem
  by decoupling the learning of representation and classification,''
  \emph{IEEE/ACM Transactions on Audio, Speech, and Language Processing},
  vol.~30, pp. 340--354, 2022.

\bibitem{Huang2022SentenceSelectLL}
W.~R. Huang, C.~Peyser, T.~N. Sainath, R.~Pang, T.~Strohman, and S.~Kumar,
  ``Sentence-select: Large-scale language model data selection for rare-word
  speech recognition,'' \emph{ArXiv}, vol. abs/2203.05008, 2022.

\bibitem{Huang2021LookupTableRL}
W.~R. Huang, T.~N. Sainath, C.~Peyser, S.~Kumar, D.~Rybach, and T.~Strohman,
  ``Lookup-table recurrent language models for long tail speech recognition,''
  \emph{ArXiv}, vol. abs/2104.04552, 2021.

\bibitem{yang2021multi}
C.-H.~H. Yang, L.~Liu, A.~Gandhe, Y.~Gu, A.~Raju, D.~Filimonov, and I.~Bulyko,
  ``Multi-task language modeling for improving speech recognition of rare
  words,'' in \emph{2021 IEEE Automatic Speech Recognition and Understanding
  Workshop (ASRU)}.\hskip 1em plus 0.5em minus 0.4em\relax IEEE, 2021, pp.
  1087--1093.

\bibitem{devlin-etal-2019-bert}
\BIBentryALTinterwordspacing
J.~Devlin, M.-W. Chang, K.~Lee, and K.~Toutanova, ``{BERT}: Pre-training of
  deep bidirectional transformers for language understanding,'' in
  \emph{Proceedings of the 2019 Conference of the North {A}merican Chapter of
  the Association for Computational Linguistics: Human Language Technologies,
  Volume 1 (Long and Short Papers)}.\hskip 1em plus 0.5em minus 0.4em\relax
  Minneapolis, Minnesota: Association for Computational Linguistics, Jun. 2019,
  pp. 4171--4186. [Online]. Available: \url{https://aclanthology.org/N19-1423}
\BIBentrySTDinterwordspacing

\bibitem{Irie2019LanguageMW}
K.~Irie, A.~Zeyer, R.~Schl{\"u}ter, and H.~Ney, ``Language modeling with deep
  transformers,'' in \emph{INTERSPEECH}, 2019.

\bibitem{taking_notes}
\BIBentryALTinterwordspacing
Q.~Wu, C.~Xing, Y.~Li, G.~Ke, D.~He, and T.-Y. Liu, ``Taking notes on the fly
  helps language pre-training,'' in \emph{ICLR}, 2021. [Online]. Available:
  \url{https://openreview.net/forum?id=lU5Rs_wCweN}
\BIBentrySTDinterwordspacing

\bibitem{Meng2021InternalLM}
Z.~Meng, N.~Kanda, Y.~Gaur, S.~Parthasarathy, E.~Sun, L.~Lu, X.~Chen, J.~Li,
  and Y.~Gong, ``Internal language model training for domain-adaptive
  end-to-end speech recognition,'' \emph{ICASSP 2021 - 2021 IEEE International
  Conference on Acoustics, Speech and Signal Processing (ICASSP)}, pp.
  7338--7342, 2021.

\bibitem{7178964}
V.~Panayotov, G.~Chen, D.~Povey, and S.~Khudanpur, ``Librispeech: An asr corpus
  based on public domain audio books,'' in \emph{2015 IEEE International
  Conference on Acoustics, Speech and Signal Processing (ICASSP)}, 2015, pp.
  5206--5210.

\bibitem{raecompressive2019}
\BIBentryALTinterwordspacing
J.~W. Rae, A.~Potapenko, S.~M. Jayakumar, C.~Hillier, and T.~P. Lillicrap,
  ``Compressive transformers for long-range sequence modelling,'' \emph{arXiv
  preprint}, 2019. [Online]. Available: \url{https://arxiv.org/abs/1911.05507}
\BIBentrySTDinterwordspacing

\bibitem{kudo-2018-subword}
\BIBentryALTinterwordspacing
T.~Kudo, ``Subword regularization: Improving neural network translation models
  with multiple subword candidates,'' in \emph{Proceedings of the 56th Annual
  Meeting of the Association for Computational Linguistics (Volume 1: Long
  Papers)}.\hskip 1em plus 0.5em minus 0.4em\relax Melbourne, Australia:
  Association for Computational Linguistics, Jul. 2018, pp. 66--75. [Online].
  Available: \url{https://aclanthology.org/P18-1007}
\BIBentrySTDinterwordspacing

\bibitem{watanabe18_interspeech}
S.~Watanabe, T.~Hori, S.~Karita, T.~Hayashi, J.~Nishitoba, Y.~Unno, N.~{Enrique
  Yalta Soplin}, J.~Heymann, M.~Wiesner, N.~Chen, A.~Renduchintala, and
  T.~Ochiai, ``{ESPnet: End-to-End Speech Processing Toolkit},'' in \emph{Proc.
  Interspeech 2018}, 2018, pp. 2207--2211.

\bibitem{inproceedings}
A.~Gulati, J.~Qin, C.-C. Chiu, N.~Parmar, Y.~Zhang, J.~Yu, W.~Han, S.~Wang,
  Z.~Zhang, Y.~Wu, and R.~Pang, ``Conformer: Convolution-augmented transformer
  for speech recognition,'' 10 2020, pp. 5036--5040.

\bibitem{feng-etal-2022-learn}
\BIBentryALTinterwordspacing
Y.~Feng, F.~Li, Z.~Song, B.~Zheng, and P.~Koehn, ``Learn to remember:
  Transformer with recurrent memory for document-level machine translation,''
  in \emph{Findings of the Association for Computational Linguistics: NAACL
  2022}.\hskip 1em plus 0.5em minus 0.4em\relax Seattle, United States:
  Association for Computational Linguistics, Jul. 2022, pp. 1409--1420.
  [Online]. Available: \url{https://aclanthology.org/2022.findings-naacl.105}
\BIBentrySTDinterwordspacing

\bibitem{ancona2018towards}
\BIBentryALTinterwordspacing
M.~Ancona, E.~Ceolini, C.~Öztireli, and M.~Gross, ``Towards better
  understanding of gradient-based attribution methods for deep neural
  networks,'' in \emph{International Conference on Learning Representations},
  2018. [Online]. Available: \url{https://openreview.net/forum?id=Sy21R9JAW}
\BIBentrySTDinterwordspacing

\bibitem{https://doi.org/10.48550/arxiv.2202.07856}
\BIBentryALTinterwordspacing
G.~Qin, Y.~Feng, and B.~Van~Durme, ``The nlp task effectiveness of long-range
  transformers,'' 2022. [Online]. Available:
  \url{https://arxiv.org/abs/2202.07856}
\BIBentrySTDinterwordspacing

\end{thebibliography}
\end{small}

\end{document}